\documentclass[a4paper,conference]{IEEEtran}
\IEEEoverridecommandlockouts

\usepackage{graphicx}
\usepackage{times}
\usepackage{mathtools}
\usepackage{epsfig}
\usepackage{graphicx}
\usepackage{amsmath}
\usepackage{amssymb}
\usepackage{makecell}
\usepackage{color, colortbl}
\definecolor{LightCyan}{rgb}{0.88,1,1}
\usepackage[shortlabels]{enumitem}
\usepackage{siunitx}

\usepackage{tablefootnote}
\usepackage{graphicx}
\ifCLASSINFOpdf
\else
\fi
\begin{document}
% \title{\LARGE End-to-end Multi-modal Multi-tasking Vehicle Control \\
% for Self-Driving Cars with Visual Perceptions}
\title{End-to-end Multi-Modal Multi-Task Vehicle Control \\
for Self-Driving Cars with Visual Perceptions}

% \author{\IEEEauthorblockN{Michael Shell}
% \IEEEauthorblockA{School of Electrical and\\Computer Engineering\\
% Georgia Institute of Technology\\
% Atlanta, Georgia 30332--0250\\
% Email: http://www.michaelshell.org/contact.html}
% \and
% \IEEEauthorblockN{Homer Simpson}
% \IEEEauthorblockA{Twentieth Century Fox\\
% Email: homer@thesimpsons.com}
% \and
% \IEEEauthorblockN{James Kirk\\ and Montgomery Scott}
% \IEEEauthorblockA{Starfleet Academy\\
% San Francisco, California 96678--2391\\
% Telephone: (800) 555--1212\\
\author{\IEEEauthorblockN{Zhengyuan Yang$^1$\textsuperscript{*}, Yixuan Zhang$^1$\textsuperscript{*}\thanks{\textsuperscript{*} Both authors contributed equally to this work.}, Jerry Yu$^2$, Junjie Cai$^2$ and Jiebo Luo$^1$}
\IEEEauthorblockA{$^1$Department of Computer Science,
University of Rochester, Rochester NY 14627, USA}
\IEEEauthorblockA{$^2$SAIC USA Innovation Center,
San Jose, CA 95134, USA}
\IEEEauthorblockA{$^1$Email: \{zyang39, jluo\}@cs.rochester.edu, \{yzh215\}@ur.rochester.edu}
\IEEEauthorblockA{$^2$Email: \{jyu, jcai\}@saicusa.com}
}
\maketitle

\begin{abstract}
Convolutional Neural Networks (CNN) have been successfully applied to autonomous driving tasks, many in an end-to-end manner. Previous end-to-end steering control methods take an image or an image sequence as the input and directly predict the steering angle with CNN. Although single task learning on steering angles has reported good performances, the steering angle alone is not sufficient for vehicle control. In this work, we propose a multi-task learning framework to predict the steering angle and speed control simultaneously in an end-to-end manner. Since it is nontrivial to predict accurate speed values with only visual inputs, we first propose a network to predict discrete speed commands and steering angles with image sequences. Moreover, we propose a multi-modal multi-task network to predict speed values and steering angles by taking previous feedback speeds and visual recordings as inputs. Experiments are conducted on the public Udacity dataset and a newly collected SAIC dataset. Results show that the proposed model predicts steering angles and speed values accurately. Furthermore, we improve the failure data synthesis methods to solve the problem of error accumulation in real road tests. %Real car tests are deployed and evaluated.

\end{abstract}
\IEEEpeerreviewmaketitle

\section{Introduction}
In many traditional self-driving car solutions \cite{chen2015deepdriving,huval2015empirical,gurghian2016deeplanes,geiger20143d,zhang2013understanding}, vehicle controls are rule based where perception and vehicle control are two individual modules. Nvidia \cite{bojarski2016end} is the first to address the task of end-to-end steering angle control, where Convolutional Neural Networks (CNN) are used to regress steering angles directly from raw pixels recorded by front-view cameras. Xu et al. \cite{xu2016end} further propose to predict the steering angle and understand the scene simultaneously in an end-to-end fashion with an FCN-LSTM architecture. 
%Despite the good steering predictions, understanding the network is also important. 
A visual attention network \cite{kim2017interpretable} is proposed to help interpret the predictions with attention heatmaps. Other approaches \cite{zeiler2014visualizing,bojarski2017explaining}  are proposed to visualize the intermediate results in CNN.

Despite the fact that the end-to-end steering angle control has achieved good results and has been well interpreted, the steering angle alone is not sufficient for vehicle control. The lack of speed commands greatly limits the potential applications of the end-to-end methods. In this work, we propose to predict the steering angle and speed command simultaneously with a multi-task learning approach. Intuitively, it is challenging to predict an accurate speed value with only visual inputs. A correct turning angle can be predicted with sufficient training data on the road, since there is only one correct way to keep the vehicle on the road. However, the driving speed is determined by a number of other factors including driver's driving habits, surrounding traffic conditions, road conditions and so on. Many factors cannot be reflected solely through front-view cameras. Therefore, we start with an easier task of discrete speed command prediction. The task is to predict discrete speed control commands of accelerating, decelerating and maintaining speed. The discrete speed control commands can be adequately inferred from front-view cameras. For example, a decelerating command is predicted when there are obstacles in the front, and an accelerating command may be predicted when the road is clear and the vehicle speed is low.

Although discrete speed commands provide a preliminary version of vehicle speed control, there exist two shortcomings. First, the levels of accelerating and decelerating are pre-fixed, which limit the smoothness \cite{rajamani2011vehicle} of the vehicle control. Second, using only the visual inputs limits the command prediction accuracy under certain circumstances. For example, when the vehicle is already fast enough or at the speed limit, the accelerating command should not be made even if the road is clear. In the initial model, the speed is inferred automatically from the input image sequences, and the prediction may be inaccurate. To achieve a better vehicle control, we propose to take previous feedback speeds as an extra modality, and predict speeds and steering angles simultaneously. The proposed model is evaluated on the public Udacity dataset \cite{udacity} and the newly collected SAIC dataset. Experiment results show that the multi-modal multi-task network provides an accurate speed prediction while further improves the state-of-the-art steering angle prediction. Furthermore, we conduct real car tests on roads similar to the SAIC dataset's testing data. We also improve the failure case data synthesis methods to solve the problem of error accumulation. 
%Details are presented in the experiment part.
% and the problem of error-accumulation are also discussed in the experiment part. 

Our main contributions include the following:
\begin{itemize}[nosep]
\item We propose a multi-modal multi-task network for end-to-end steering angle and speed prediction.
\item We collect a new SAIC dataset containing the driving records during the day and night. The dataset will be released upon the publication of this work.
\item 
% We propose a number of pre-processing methods to alleviate noise in real car tests. 
We improve the failure case data synthesis methods to solve the problem of error accumulation in real car tests.
\end{itemize}

\section{Related Work}
% One of the early computer vision based self-driving car work is the DARPA Autonomous Vehicle (DAVE) \cite{lecun2004dave}, developed as a Defense Advanced Research Projects Agency (DARPA) seedling project. 
ALVINN \cite{pomerleau1989alvinn} is one of the earliest successful neural network based self-driving vehicle project. The network is simple and shallow, but it manages to do well on simple roads with a few obstacles.
% DAVE-2 was inspired by the pioneering work of Pomerleau \cite{pomerleau1989alvinn} who in 1989 built the Autonomous Land Vehicle in a Neural Network (ALVINN) system
With the development of deep learning \cite{lecun2015deep,krizhevsky2012imagenet}, many systems use CNN for environment perception and steering angle prediction. Nvidia is the first to adopt Convolutional Neural Networks (CNN) for end-to-end steering angle prediction \cite{bojarski2016end}. They propose to predict steering angles with only three front-view cameras and manage to control the vehicle with the proposed system. There exist three main approaches: behavior reflex CNN, mediated perception and privileged training. Behavior reflex CNN \cite{bojarski2016end,kim2017interpretable,bojarski2017explaining,chi2017deep,bojarski2016visualbackprop,muller2006off} directly predict the steering angle from the visual inputs. The system has a low model complexity and can be robust with enough training data. Furthermore, it has a good generalization ability. However, the performance is limited in complicated environments and the results are difficult to interpret. Some systems propose visualization methods \cite{zeiler2014visualizing,bojarski2017explaining} and include attention mechanisms \cite{kim2017interpretable,xu2015show,chen2017brain} to better interpret the results. Mediated perception \cite{chen2015deepdriving} first maps visual inputs into several pre-defined parameters to depict the surroundings. Rule based methods then produce control commends with the estimated parameters. Such methods have a better vehicle control smoothness \cite{rajamani2011vehicle} but can only work in limited scenarios. Designing ideal control rules is also difficult. Privileged training \cite{xu2016end,sharmanska2013learning} is a multi-task approach that understands the scene and predicts vehicle commands simultaneously. The main limitation is the large amount of training data required. In this work, we expand the behavior reflex CNN with a multi-modal multi-task framework. Feedback speeds are used as an extra modality for steering angle and speed prediction.

\section{Method}
\begin{figure}[t]
\begin{center}
   \centerline{\includegraphics[width=9cm]{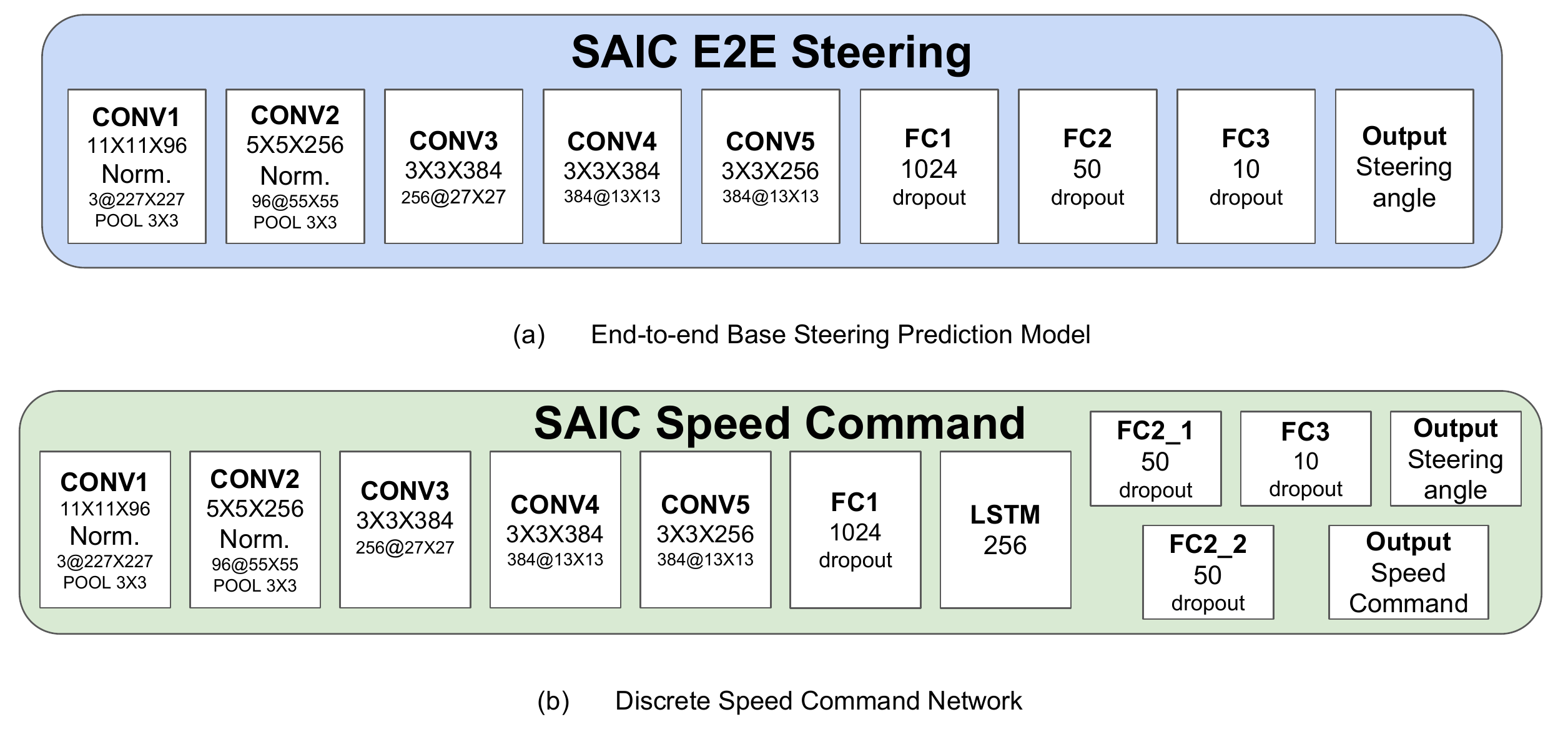}}
\end{center}
\vspace{-0.2in}
\caption{End-to-end steering and discrete speed command model.}
\vspace{-0.1in}
\label{fig:base_mdoel}
\end{figure}

% To achieve basic speed control besides steering angle prediction, we propose the multi-tasking model to predict speed command and steering angle simultaneously. In this section, we first introduce the pre-processing methods used on real-car experiment. The ConvNet, Inertial Model and multi-tasking Model are then introduced.

In this section, we first introduce the base CNN model for end-to-end steering angle prediction. Based on the improved CNN structure, a multi-task network is proposed to predict the steering angle and discrete speed command simultaneously by taking an image sequence as the input. Finally, we propose a multi-modal multi-task network that takes previous feedback speeds as an extra modality and predicts the speed and steering angle simultaneously.
\subsection{Base Steering Model} \label{base steering model}
% Inspired by\cite{bojarski2016end}, the convolutional network has an outstanding performance of extracting visual features from images. We construct a base ConvNet model, shown in Fig 1., including Base ConvNet and steering control parts. Our network consists of 13 layers, including 5 convolutional layers, 2 normalization layers, 2 maxpooling layers, 3 fully connected layer and 2 dropout layer.

% The input of ConvNet includes processed images in HSV format, speed, angle in radius and speed command. Five convolutional layers are designed for feature extraction, as discussed  by Bojarski et al. \cite{bojarski2016end} After convolutional layers, we follow up with three fully connected layers, in shape of 1024, 50, 1 respectively. The first two fully connected layers lead to the prediction of steering angle. The last fully connected layer serve as the output of steering angle. Normalization layers allows the network to normalize data stream within the training process, and also accelerate GPU computation. Maxpooling layers can prevent spatial locational information loss, when the strongest activation propagates through the network, mentioned by Lee et al. \cite{lee2009convolutional} Both normalization layer and pooling layer are hard-coded, and do not involved in back-propagation. Dropout layer is utilized for avoiding overfitting, because of abundant volume of our dataset. \cite{srivastava2014dropout}

It is shown in \cite{bojarski2016end} that CNN has a good ability in extracting visual features and is capable of directly regressing the steering angle from raw pixels. Inspired by previous end-to-end steering angle prediction systems, we propose an improved CNN structure for this task with two improvements. As shown in Figure \ref{fig:base_mdoel} (a), the model consists of 9 layers including 5 convolutional layers and 4 fully connected layers. Unlike previous work \cite{bojarski2016end}, the convolutional layers are designed based on AlexNet \cite{krizhevsky2012imagenet,jia2014caffe} and a large kernel size is adopted in the first few layers. Experiments show that larger kernels are suitable for front-view cameras and can better capture the environment features. Another improvement is changing the aspect ratio of the input image to 1:1. Previous methods \cite{bojarski2016end,kim2017interpretable} resize the input with a fixed aspect ratio of around 2.5:1. The convolutional kernels with a same width and height are then adopted. According to human intuitions though, visual content distributed along the y-axis is more informative for steering angle prediction. This implies that CNN kernels should have a larger width than height. For simplicity, we squeeze the input images in width to an aspect ratio of 1:1 and continue using the square kernels. Experiments show that the two improvements, the larger kernel size and reshaped aspect ratio, improve the performance of the end-to-end steering angle prediction. We further combine these two improvements with larger networks like VGG \cite{simonyan2014very} and ResNet \cite{he2016deep}. Although the model tends to overfit on all the evaluated datasets, the combination is promising in the future when larger datasets are available. 

The mean absolute error is adopted as the training loss function. In addition, We apply different loss weights to alleviate the problem of data imbalance, as going straight appears more frequently than turning. The data with a small steering angle has a small training loss weight and the turning data has a larger weight. This technique is applied to all steering angle prediction models in this paper.

\subsection{Discrete Speed Command Network} \label{speed command}
The end-to-end steering angle control successfully proves the feasibility of generating vehicle controls directly from front view cameras. However, the steering angle alone is not sufficient for vehicle control. The speed is another important parameter that needs to be predicted. Unlike the steering angle though, predicting the vehicle speed solely from a front view camera is counterintuitive, because even human drivers drive at different speeds given a similar road condition. Therefore, it is more reasonable to predict the speed control command from visual information, instead of directly predicting the desired speed values. For example, all drivers should slow down when the vehicle is too close to other cars or obstacles, and most drivers speed up when the road is clear. Based on this observation, we first propose a multi-task framework that predicts discrete speed commands and steering angles simultaneously. The model is called the speed command network.

As shown in Figure \ref{fig:base_mdoel} (b), the speed command network takes an image sequence as the input and predicts discrete speed commands and steering angles simultaneously. The convolutional layers have a same structure as in the base steering model. The encoded visual features are fed into an LSTM layer for temporal analysis. The output image sequence feature is used for both steering angle regression and speed command classification. As a first step, the speed commands contain three classes: "accelerating", "decelerating" and "maintaining speed". The cross entropy loss is used for speed command classification and the mean absolute error is calculated for steering angle prediction. A weighting term is added as a hyper-parameter to adjust the importance of the two tasks.
\subsection{Multi-modal Multi-task Network}
\begin{figure}[t]
\begin{center}
   \centerline{\includegraphics[width=8cm]{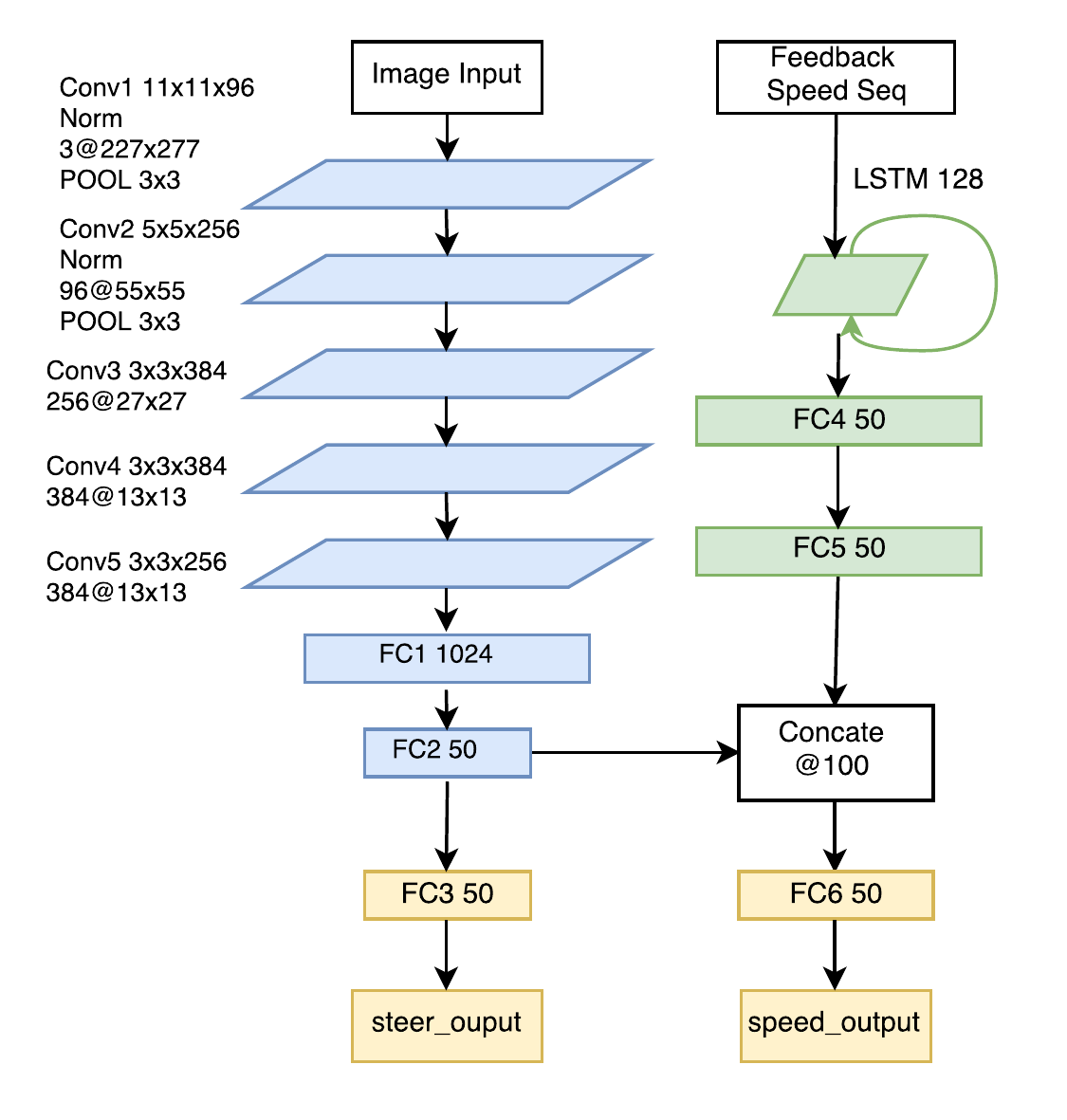}}
\end{center}
\vspace{-0.2in}
\caption{End-to-end multi-modal multi-task vehicle control model. Different colors represent different modules.}
\vspace{-0.1in}
\label{fig:model}
\end{figure}
% To achieve a better vehicle control, we propose to take previous speed feedbacks as an extra modality and predict speed and steering angle simultaneously.
The speed command network provides an initial framework for vehicle speed control. However, the performance is limited due to the lack of input information. The visual contents from the front view cameras alone are not sufficient for accurate speed command prediction. For example, in most cases it is reasonable to speed up when the road is clear, but it is not the case when the vehicle is already at a high speed. Similarly, there is no need to slow down when the vehicle is already slow enough. These failure cases are observed in the experiments and vehicle speeds are necessary for making a good speed command prediction. Theoretically, the vehicle speed can be predicted from image sequences, but the prediction is difficult and inaccurate. A more reasonable solution is to directly adopt the feedback speeds. Therefore, we propose a multi-modal multi-task network to predict the values of steering angles and speeds simultaneously by taking previous feedback speeds as an extra modality.

The model structure is shown in Figure \ref{fig:model}. The network contains a visual encoder and a speed encoder. The visual encoder takes only one frame as inputs instead of using the CNN + LSTM structure. This greatly reduces the amount of computation, therefore guarantees a high FPS and a real-time performance even with low performance GPUs. The speed encoder encodes the pattern of previous feedback speed sequences. The encoded visual features are used for steering angle prediction, and the concatenation of visual features and feedback speed features are adopted for speed prediction. Both steering angle prediction and speed prediction apply mean absolute loss as a loss function, and a weighting parameter is tuned to adjust the weight between the two loss terms.

\section{Dataset}
\begin{figure*}[t]
\begin{center}
   \centerline{\includegraphics[width=17cm]{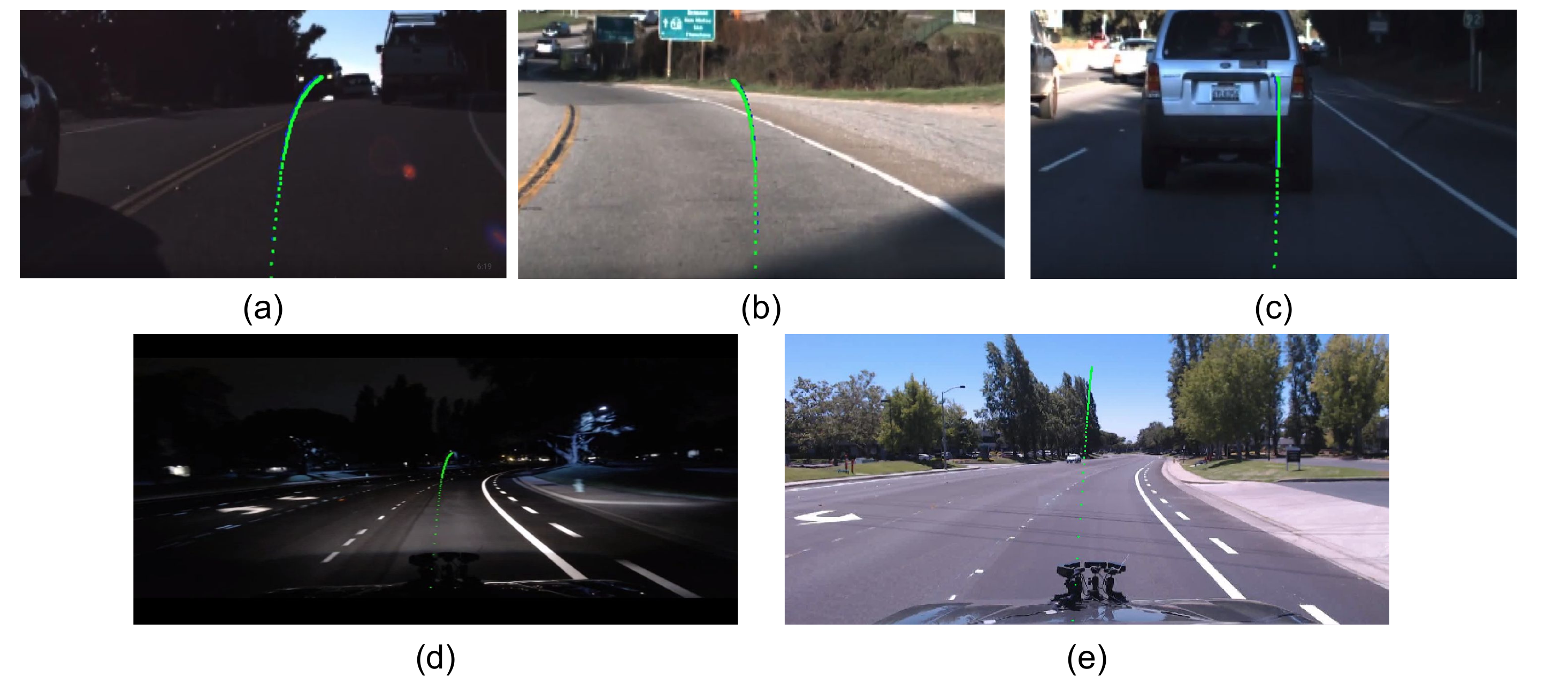}}
\end{center}
\vspace{-0.3in}
\caption{Example frames and predictions on the Udacity and SAIC datasets. First row: the Udacity dataset. Second row: the SAIC dataset.}
\vspace{-0.1in}
\label{fig:data_example}
\end{figure*}

% The proposed framework is first evaluated on several public dataset. Furthermore, the framework is also tested on our own collected data. The steering angle only model is selected as the baseline. Experiments show the multi-tasking setting both provide the extra speed command and improve the accuracy of steering angle prediction.
In this section, we first introduce the public Udacity dataset \cite{udacity}. The collection and statistics of the SAIC dataset is then discussed. Example frames of both datasets are shown in Figure \ref{fig:data_example}. Finally, we introduce the data pre-processing methods. 
% Add statistics to SAIC dataset.
\subsection{Dataset}
\subsubsection{Udacity}
The Udacity dataset \cite{udacity} is originally provided for an online challenge. The dataset contains six video clips with a total duration of around 20 minutes. Speed values, steering angles and video streams from three front view cameras are recorded.
\subsubsection{SAIC} \label{saic}
In order to obtain a larger data size and find regions for real road test, we record and build the SAIC dataset. The dataset includes five hours of driving data in north San Jose area, mostly on urban roads. The dataset contains the driving data in both day and night. The vehicle goes between several nodes and each trip between the nodes has a duration of around ten minutes. Parking, waiting at traffic lights and some other conditions are considered as noisy parts and filtered out. After filtering out the noisy videos, two hours' data is split into training, validation and testing set. A whole video of a certain trip between two nodes is atomic in set splits. Three drivers are included to avoid biasing towards a specific driving behavior. Similarly, video streams, speed values and steering angles are recorded. The video streams contain videos from one center and two side front view cameras with a frame rate of 30 frames per second.
\subsection{Data Pre-Processing} \label{preprocessing}
\subsubsection{Image Pre-Processing}
We adopt several image pre-processing and data augmentation techniques to improve the robustness and prediction accuracy of the proposed system. The robustness under various lighting conditions is a major challenge for camera-based systems. We show that converting frames into different color spaces can improve the robustness towards lighting changes. The input frames are converted from RGB color space to HSV. A small rotation angle is randomly added to simulate the camera vibrations on vehicles. For data augmentation, random horizontal flips are first adopted. Another important technique is data synthesis with side cameras, which generates simulated failure cases for training.
\subsubsection{Speed Command Generating}
% Since our objective is to make a classification of three driving conditions of Speed up, Slow down and No change, we generate labels of speed command from both our own dataset and public datasets. For some public dataset, such as Comma.ai which have already provided with acceleration at each frame, we simply convert the numerical acceleration into three specific speed commands, "1", "-1", "0", respectively represent speed up, slow down and no change. For our own dataset, SAIC, there is no given acceleration data, but speed is given. Therefore, we generate the labels of speed command on our own. In detail, for those dataset given acceleration, we simply set a threshold to distinguish the boundary of three speed conditions. After a considerable visual judgment, we choose 0.25m/$s{^2}$ and -0.25m/$s{^2}$ to be the upper boundary and lower boundary respectively. In this way, any acceleration larger than 0.25m/$s{^2}$ would be considered as "Speed up", and any acceleration smaller than -0.25m/$s{^2}$ would be considered as "Slow down". And this standard applies on all the dataset we use in our experiment.
We introduce the methods for generating discrete speed commands. We first calculate acceleration from speed sequences with the following equation:
\begin{equation}
acce = \frac{speed_e - speed_s}{interval}
\end{equation}
where $acce$ is the calculated acceleration, $speed_e$ is the speed at the end of the interval, $speed_s$ is the speed at the start of the interval. The $interval$ is set to one second in our experiment. Two acceleration thresholds are then selected to generate the labels for the three classes: "accelerating", "decelerating" and "maintaining speed". According to manual visual observations and domain experts' suggestions, 0.25m/$s{^2}$ and -0.25m/$s{^2}$ are selected as the upper and lower thresholds, respectively. The accelerations larger than 0.25m/$s{^2}$ are labeled as "Accelerating", and the values smaller than -0.25m/$s{^2}$ is tagged with "Decelerating". Remaining minor speed changes are labeled as "Maintaining Speed". 
% For our own dataset, we consider that human would have a short amount of time to take action, and usually it is between 0.7 and 3 seconds, and some accident reconstruction specialists use 1.5 seconds. Thus, we calculate the acceleration at the time interval of 1 second. Concretely speaking, we use the speed of 20th frame to minus the speed of 1st frame at each second (No need to divided by time period, mathematically divided by 1 does not affect the calculation). 

\section{Experiment}
The proposed method is evaluated on the public Udacity dataset \cite{udacity} and the collected SAIC dataset. We first present the results of steering angle prediction. The performances of speed command predictions and speed value estimations are then evaluated. Finally, we introduce real car tests and an improved data synthesis method that solves the error accumulation problem in vehicle tests.

% We could add some visualizations, if space is enough. Like in \cite{chi2017deep}.

\begin{table}[!t]
\caption{Experiment results of steering angle prediction on Udacity}
\label{table:angle}
\centering
\begin{tabular}{|c|c|c|}
\hline
Method & Angle (MAE in $degree$) \\ 
\hline
Nvidia's PilotNet \cite{bojarski2016end} & 4.26 \\
\hline
Cg Network \cite{cg23}  & 4.18\\
% \tablefootnote{One of community models in Udacity Self-Driving challenge}
\hline
Base Steering Model & 2.84 \\
\hline
Discrete Speed Command Network & 1.85 \\
\hline
Multi-modal Multi-task Network & 1.26 \\
\hline
\end{tabular}
\end{table}

% \begin{table}[!t]
% \caption{Experiment results of steering angle prediction on Udacity dataset}
% \label{table:angle}
% \centering
% \begin{tabular}{|c|c|c|}
% \hline
% \multicolumn{3}{|c|}{Angle (MAE in $degree$)}\\
% \hline
% Method & Udacity & SAIC \\ 
% \hline
% Nvidia's PilotNet \cite{bojarski2016end} & 4.260 & \\
% \hline
% Cg Network \tablefootnote{One of community models in Udacity Self-Driving challenge} & 4.175 & \\
% \hline
% Base Steering Model & 2.844 & \\
% \hline
% Multi-modal Multi-task Network & 1.255 & \\
% \hline
% \end{tabular}
% \end{table}

\begin{table}[!t]
\caption{Results of speed value prediction on the Udacity dataset and the SAIC dataset with Multi-modal Multi-task Network}
\label{table:speed}
\centering
\begin{tabular}{|c|c|}
\hline
Dataset & Speed (MAE in $m/s$) \\ 
\hline
Udacity \cite{udacity} & 0.19 \\
\hline
SAIC & 0.45 \\ 
\hline
\end{tabular}
\end{table}

\subsection{Steering Angle Prediction}
% As to the steering angle prediction, we conduct five experiments by applying both existing models, for instance\cite{bojarski2016end} Cg Network, and our own models. 
We first evaluate the performance of end-to-end steering angle prediction. The proposed multi-modal multi-task model is compared with several state-of-the-art models and the proposed improved single task network. Nvidia's PilotNet\cite{bojarski2016end} and the Cg Network \cite{cg23} proposed in the Udacity Self-Driving challenge is reimplemented and selected for comparison. As a regression task, the performance is reported in terms of MAE (Mean Absolute Error) in degree. Furthermore, we discard low speed data that is slower than $4 m/s$. It is observed that steering angles tend to be much larger when vehicles are almost stopped, which are considered as noise in steerings.
% since low-speed data might result from various circumstances which are probably outside our research purpose

The models are first evaluated on the Udacity dataset. As shown in Table \ref{table:angle}, the propose model is compared to the reimplemented Nvidia's PilotNet\cite{bojarski2016end} and the Cg Network \cite{cg23} from the Udacity Self-Driving challenge. Nvidia's PilotNet has five convolutional layers and five fully connected layers with an input of $200*66$. The Cg Network is even simpler with three convolutional layers and two fully connected layers. Furthermore, the proposed base steering model and the speed command network are compared in order to protrude the advantage of the proposed Multi-modal Multi-task network. 
% we also experiment on a base steering model(detail in Section\ref{base steering model}). This model architecture is part of Multi-task network, and focus on steering angle prediction. 
% Although the network structures of PilotNet and our base steering model are inspired by AlexNet\cite{krizhevsky2012imagenet}, major difference is that we have a more concrete and considerable data pre-processing, as mentioned in Section \ref{preprocessing}. 
% As to our final proposed model, Multi-modal Multi-task Network, shown in Figure \ref{fig:model}, not only have images, but also previous feedback speed as input, compared with base steering model. 
% In Table \ref{table:angle}, we compare the prediction performance with four models. Encouragingly, the performance of our Base Steering model has already exceeded PilotNet and Cg Network. As mentioned previously, our considerable data pre-processing significantly improves the performance of steering angle prediction, even with a similar model. Our final proposed model, Multi-task model even outperforms over base steering model. 

As shown in Table \ref{table:angle}, the improved base steering mode outperforms the reimplemented Nvidia's PilotNet\cite{bojarski2016end} and the Cg Network \cite{cg23}. This proves the effectiveness of the proposed CNN structure with larger kernel sizes and adjusted aspect ratios. PilotNet is proposed to work on other unpublished datasets, which might limit its performance in our evaluations.

By comparing the multi-task speed command model to the base steering model, we observe a further improvement in the steering accuracy from \ang{2.84} to \ang{1.85}. This shows that the multi-task model provides additional speed prediction while further improves the performance of the steering angle prediction task. The multi-modal multi-task model further improves the steering accuracy from \ang{1.85} to \ang{1.26}. As an extension, the multi-modal multi-task model takes previous feedback speeds as an extra modality of inputs and predict the speed and steering angle simultaneously. The extra modality and task help the model better understand the vehicle condition and thus generate a more accurate steering angle prediction.

% The reason is that Multi-task model has extra input information, previous feedback speed, and simultaneously it is making prediction of speed of next frame. This provides the steering angle prediction with a reasonable scenario, where people's driving is depended on both surroundings and how fast they go. For instance, with a high speed, the inertial effect would cause steering control much harder than a regular driving speed condition. Our training data is from human drivers, therefore the model should take these possible situations into consideration. Our experiment result proves our idea, where the performance of Multi-task model has improved about 70\% over PilotNet, 69\% over Cg Network, 56\% over Base Steering model. 

Furthermore, we apply single exponential smoothing with thresholds \cite{hyndman2008forecasting,kim2017interpretable} on the final steering angle output. The intuition is to improve the vehicle control smoothness. The smoothing process adopts the following equation:
\begin{equation}
% angle_f = \alpha * angle_p + (1 - \alpha) * angle_b
% UPDATE EQUATION
\hat{\theta}_t = \alpha * \theta_t + (1 - \alpha) * \hat{\theta}_{t-1}
\end{equation}
where $\hat{\theta}_t$ is the smoothed steering angle output at the current frame, $\theta_t$ is the steering angle prediction at the current frame and $\hat{\theta}_{t-1}$ is the smoothed steering angle at the last timestamp. $\alpha$ is the smoothing factor and is set to 0.2.

Experiments are also conducted on the newly collected SAIC dataset. We achieve a steering angle prediction accuracy of \ang{0.17} with the multi-modal multi-task network.
% One of the most common problem of steering angle prediction in Autonomous Driving is error-accumulation. The prediction generated with both current and past steering angle, applying a proper weight, can effectively reduce the influence of error-accumulation. As Figure \ref{table:angle} shown, although the predicted steering angle is fluctuate around the ground truth, it is maintained in a reasonable range. Based on our experiment(On Udacity and SAIC dataset), we find that when $\alpha$ is 0.2, the performance is the best. 

\subsection{Discrete Speed Command Prediction}
\begin{figure}[t]
\begin{center}
   \centerline{\includegraphics[width=10cm,height=7cm]{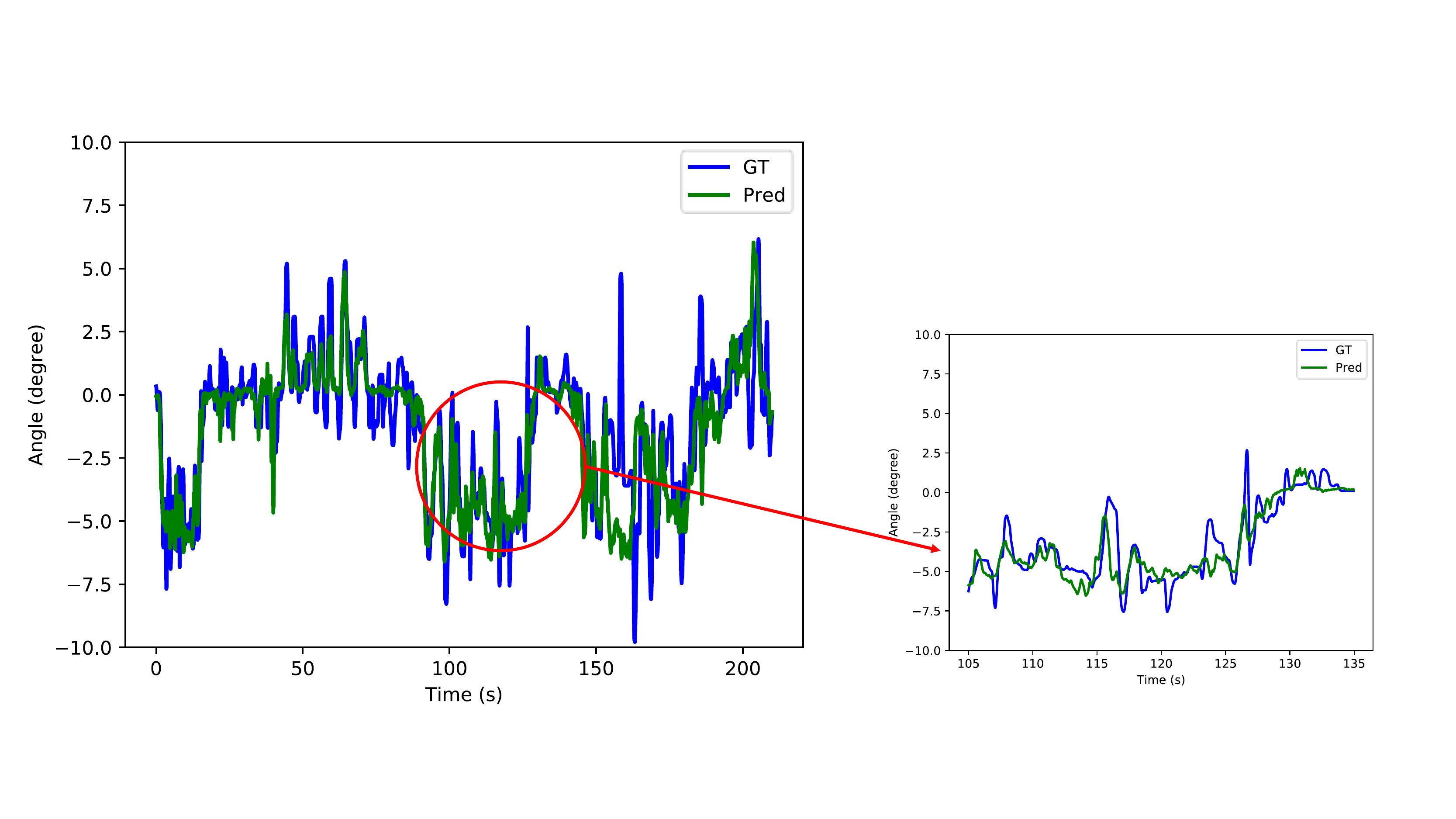}}
\end{center}
\vspace{-0.6in}
\caption{Steering angle prediction results by the multi-modal multi-task network on the Udaicty dataset \cite{udacity}.}
\vspace{-0.1in}
\label{fig:angle}
\end{figure}

As introduced in Section \ref{speed command}, we first simplify the speed prediction problem into a multi-class classification problem where the classes are discrete speed commands. Experiments are conducted on the Udacity dataset and the SAIC dataset with the model structure shown in Figure \ref{fig:base_mdoel} (b).
% The input becomes a sequence of previous image frames. 
We convert acceleration value sequences into discrete speed command sequences containing the labels of `accelerating', `decelerating' and `maintaining speed'. All discrete command labels are transferred into one-hot vectors. 
% Although the idea of Discrete Speed Command Prediction is more intuitive, the experiment result is not as appealing as we expect. The speed command classification accuracy on Udacity dataset is about 65\%.

On the Udacity dataset, we achieve a speed command classification accuracy of 65.0\%. Furthermore, the multi-task model improves the steering angle prediction accuracy from \ang{2.84} to \ang{1.85}. Despite the improvements in steering angle prediction, the results are limited. After observing the error classes, we find two major reasons for the failure cases. First, the generated speed commands are noisy with the human factors-related speed changes. Increasing the interval in calculating the acceleration can alleviate the problem, but it leads to a delay in generating the speed command. Another problem is that it is inherently difficult to predict the speed command with only the visual inputs. As mentioned earlier, there is no need to slow down when the vehicle is already slow enough even if the obstacles are close to the vehicle. To solve these problems, we further propose the multi-modal multi-task network.
% After comprehensive analysis, we think that the poor performance results from our own handcrafted labeling, and the dataset is not generated for speed command. At some frames, the speed has a great fluctuation. For example, within one time interval, a drastic speed change can lead to more than one speed command. However, our labeling strategy will only consider one speed command class at this circumstance. Meanwhile, a very short time interval would also lead to a over-fluctuated condition of speed command. As a result, our speed command labels will include too much noise. Additionally, we explore the raw acceleration data, but it still seems not as good as we expect. From our perspective, an on-purpose generated dataset would help correcting this fault-labeling problem, and a better result may come out later. 

\subsection{Speed Control Value Prediction}
\begin{figure}[t]
\begin{center}
   \centerline{\includegraphics[width=9.5cm,height=7cm]{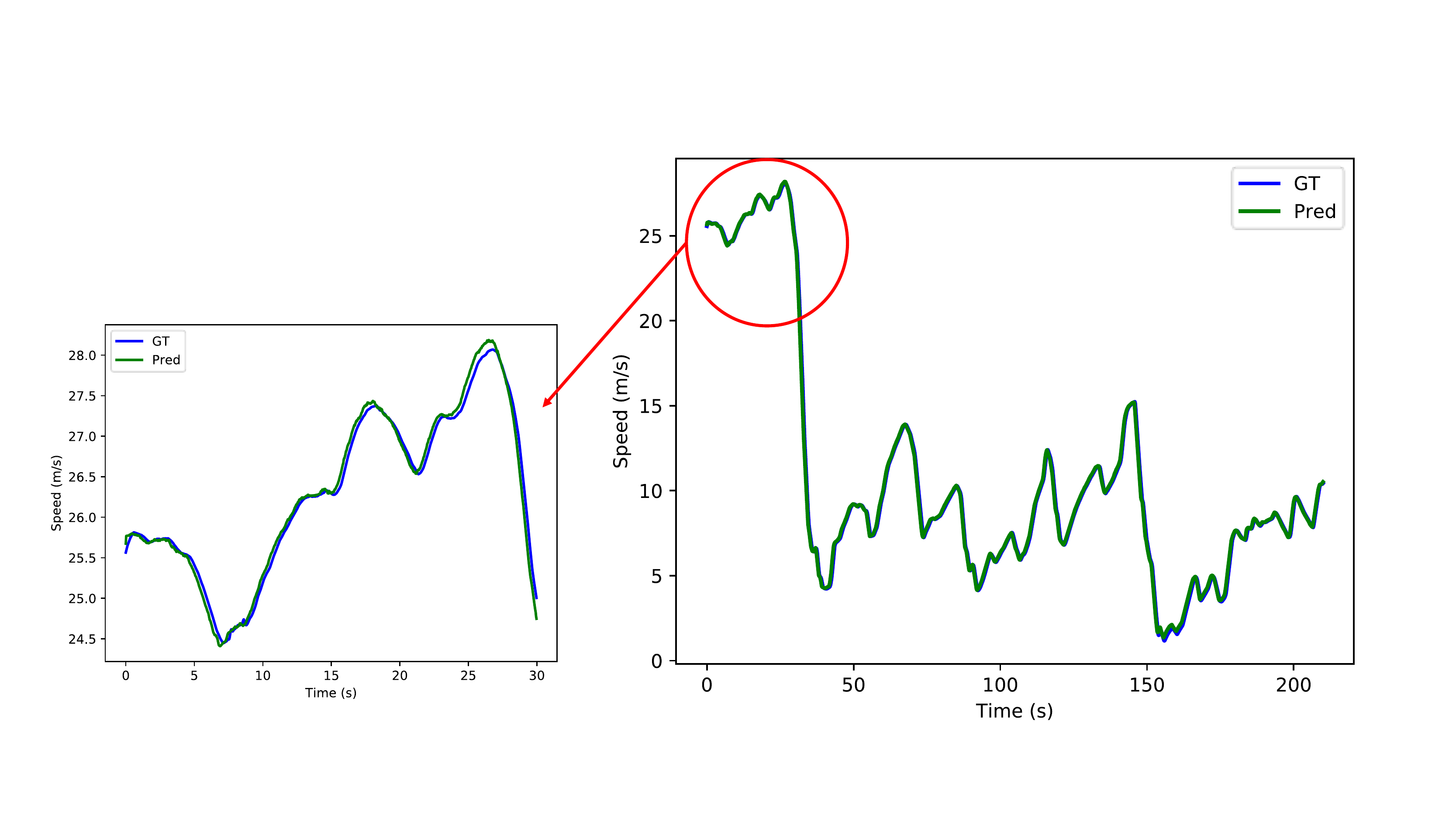}}
\end{center}
\vspace{-0.6in}
\caption{Speed value prediction results by the multi-modal multi-task network on the Udaicty dataset \cite{udacity}.}
\vspace{-0.1in}
\label{fig:speed}
\end{figure}

The multi-modal multi-task network, shown in Figure \ref{fig:model}, directly predicts the speed value of the next frame by utilizing both visual inputs and feedback speed inputs. Different from speed command prediction, the ground truth labels of speed values are numerical values in unit of $m/s$ and the problem is now modeled as a regression task. For inputs, the visual input is one single frame and the feedback speeds contain the speeds of 10 previous timestamps. Similar to steering angle prediction, the low speed data (less than $4 m/s$) is discarded to ensure a consistent driving condition. Experiments are conducted on both the Udacity and the SAIC datasets. The speed prediction performance of the multi-modal multi-task model is shown in Table \ref{table:speed}. We achieve an MAE of $0.19 m/s$ on the Udacity dataset and an MAE of $0.45 m/s$ on the SAIC dataset. Since the speed prediction task is novel, we did not find any baselines for comparison. The speed prediction results are plotted in Figure \ref{fig:speed} and the predicted values match well with the ground truth. Furthermore, an improvement in steering angle prediction is observed with the multi-modal multi-task model.
% Considering that Udacity dataset is generated from a more complex driving condition, 0.193 $m/s$ in MAE is an extremely low error. From Figure \ref{fig:speed}, we can find that the speed prediction fits the ground truth well. Considering that SAIC dataset is generated from a more smooth driving condition, 0.056 $m/s$ in MAE is reasonably lower than the MAE of Udacity dataset. However, there is no relevant publication about camera-based autonomous driving multi-task model, we cannot directly compare our performance with others. Intuitively, regardless of whether experiment performance of Udacity dataset or SAIC dataset has already become the state of the art in this certain area. 

\subsection{Road Tests and Data Synthesis}
Despite the good simulation results, we further discuss the challenges and corresponding solutions used in road tests. The major challenge in road tests is error accumulation. The accumulated error in the steering angle reflects as a shift vertical to the road and finally leads to the drift away of the vehicle. Similar error accumulation is also observed in speed control, as the feedback speeds have been used for future speed predictions. Therefore, the input data should contain adequate samples of recovering from failures. However, failure case data collection is dangerous and infeasible, since human drivers would have to frequently drive off the road and recover.

Inspired by \cite{bojarski2016end}, we use side cameras to synthesize the failure case data for steering angle prediction. An artificial recovering angle is added with the following equation:
\begin{equation}
\theta_f = \theta_r + \arctan(\frac{d_y}{s*t_r})
\end{equation}
where $\theta_f$ is the simulated steering angle with a recovering angle added, $\theta_r$ is the driver's steering angle corresponding to the center camera, $d_y$ is the distance between the side and center cameras, $s$ is the current speed and $t_r$ is the time of the whole recovering process. In our experiments, the camera offset $d_y$ is 20 inches (50.8 cm). Based on expert knowledge, we adopt a recovering time of one second in our experiments. Furthermore, we extend the data synthesis methods to speed data synthesis. Experiments on real cars show that vehicles would drift away without the data synthesis method. With the synthesized failure cases added, vehicles manage to drive autonomously on the road under a similar condition in SAIC.

\section{Conclusion}
In this paper, we address the challenging task of end-to-end vehicle control in terms of both the speed and steering angle. A multi-modal multi-task framework is proposed for the joint task. The model takes front-view camera recordings and feedback speed sequences as the input. Experiments show that the proposed multi-task framework predicts the speed value accurately and further improves the accuracy of steering angle prediction. A new SAIC dataset is collected  for evaluation and further studies. Finally, the error accumulation problem in real vehicle road tests are introduced. An extended data synthesis method is proposed for failure case simulation, which help solve the error accumulation problem.
\section*{Acknowledgments}
We thank the support of New York State through the Goergen Institute for Data Science, and SAIC USA. 

%\section*{Acknowledgment}




\bibliographystyle{IEEEtran}
\bibliography{IEEEfull.bib}

\begin{thebibliography}{10}
\providecommand{\url}[1]{#1}
\csname url@samestyle\endcsname
\providecommand{\newblock}{\relax}
\providecommand{\bibinfo}[2]{#2}
\providecommand{\BIBentrySTDinterwordspacing}{\spaceskip=0pt\relax}
\providecommand{\BIBentryALTinterwordstretchfactor}{4}
\providecommand{\BIBentryALTinterwordspacing}{\spaceskip=\fontdimen2\font plus
\BIBentryALTinterwordstretchfactor\fontdimen3\font minus
  \fontdimen4\font\relax}
\providecommand{\BIBforeignlanguage}[2]{{%
\expandafter\ifx\csname l@#1\endcsname\relax
\typeout{** WARNING: IEEEtran.bst: No hyphenation pattern has been}%
\typeout{** loaded for the language `#1'. Using the pattern for}%
\typeout{** the default language instead.}%
\else
\language=\csname l@#1\endcsname
\fi
#2}}
\providecommand{\BIBdecl}{\relax}
\BIBdecl

\bibitem{chen2015deepdriving}
C.~Chen, A.~Seff, A.~Kornhauser, and J.~Xiao, ``Deepdriving: Learning
  affordance for direct perception in autonomous driving,'' in
  \emph{Proceedings of the IEEE International Conference on Computer Vision},
  2015, pp. 2722--2730.

\bibitem{huval2015empirical}
B.~Huval, T.~Wang, S.~Tandon, J.~Kiske, W.~Song, J.~Pazhayampallil,
  M.~Andriluka, P.~Rajpurkar, T.~Migimatsu, R.~Cheng-Yue \emph{et~al.}, ``An
  empirical evaluation of deep learning on highway driving,'' \emph{arXiv
  preprint arXiv:1504.01716}, 2015.

\bibitem{gurghian2016deeplanes}
A.~Gurghian, T.~Koduri, S.~V. Bailur, K.~J. Carey, and V.~N. Murali,
  ``Deeplanes: End-to-end lane position estimation using deep neural
  networksa,'' in \emph{Proceedings of the IEEE Conference on Computer Vision
  and Pattern Recognition Workshops}, 2016, pp. 38--45.

\bibitem{geiger20143d}
A.~Geiger, M.~Lauer, C.~Wojek, C.~Stiller, and R.~Urtasun, ``3d traffic scene
  understanding from movable platforms,'' \emph{IEEE transactions on pattern
  analysis and machine intelligence}, vol.~36, no.~5, pp. 1012--1025, 2014.

\bibitem{zhang2013understanding}
H.~Zhang, A.~Geiger, and R.~Urtasun, ``Understanding high-level semantics by
  modeling traffic patterns,'' in \emph{Proceedings of the IEEE International
  Conference on Computer Vision}, 2013, pp. 3056--3063.

\bibitem{bojarski2016end}
M.~Bojarski, D.~Del~Testa, D.~Dworakowski, B.~Firner, B.~Flepp, P.~Goyal, L.~D.
  Jackel, M.~Monfort, U.~Muller, J.~Zhang \emph{et~al.}, ``End to end learning
  for self-driving cars,'' \emph{arXiv preprint arXiv:1604.07316}, 2016.

\bibitem{xu2016end}
H.~Xu, Y.~Gao, F.~Yu, and T.~Darrell, ``End-to-end learning of driving models
  from large-scale video datasets,'' \emph{arXiv preprint arXiv:1612.01079},
  2016.

\bibitem{kim2017interpretable}
J.~Kim and J.~Canny, ``Interpretable learning for self-driving cars by
  visualizing causal attention,'' \emph{arXiv preprint arXiv:1703.10631}, 2017.

\bibitem{zeiler2014visualizing}
M.~D. Zeiler and R.~Fergus, ``Visualizing and understanding convolutional
  networks,'' in \emph{European conference on computer vision}.\hskip 1em plus
  0.5em minus 0.4em\relax Springer, 2014, pp. 818--833.

\bibitem{bojarski2017explaining}
M.~Bojarski, P.~Yeres, A.~Choromanska, K.~Choromanski, B.~Firner, L.~Jackel,
  and U.~Muller, ``Explaining how a deep neural network trained with end-to-end
  learning steers a car,'' \emph{arXiv preprint arXiv:1704.07911}, 2017.

\bibitem{rajamani2011vehicle}
R.~Rajamani, \emph{Vehicle dynamics and control}.\hskip 1em plus 0.5em minus
  0.4em\relax Springer Science \& Business Media, 2011.

\bibitem{udacity}
``Udacity. public driving dataset,''
  \url{https://www.udacity.com/self-driving-car}, accessed: 2017-03-07.

\bibitem{pomerleau1989alvinn}
D.~A. Pomerleau, ``Alvinn: An autonomous land vehicle in a neural network,'' in
  \emph{Advances in neural information processing systems}, 1989, pp. 305--313.

\bibitem{lecun2015deep}
Y.~LeCun, Y.~Bengio, and G.~Hinton, ``Deep learning,'' \emph{Nature}, vol. 521,
  no. 7553, pp. 436--444, 2015.

\bibitem{krizhevsky2012imagenet}
A.~Krizhevsky, I.~Sutskever, and G.~E. Hinton, ``Imagenet classification with
  deep convolutional neural networks,'' in \emph{Advances in neural information
  processing systems}, 2012, pp. 1097--1105.

\bibitem{chi2017deep}
L.~Chi and Y.~Mu, ``Deep steering: Learning end-to-end driving model from
  spatial and temporal visual cues,'' \emph{arXiv preprint arXiv:1708.03798},
  2017.

\bibitem{bojarski2016visualbackprop}
M.~Bojarski, A.~Choromanska, K.~Choromanski, B.~Firner, L.~Jackel, U.~Muller,
  and K.~Zieba, ``Visualbackprop: visualizing cnns for autonomous driving,''
  \emph{arXiv preprint arXiv:1611.05418}, 2016.

\bibitem{muller2006off}
U.~Muller, J.~Ben, E.~Cosatto, B.~Flepp, and Y.~L. Cun, ``Off-road obstacle
  avoidance through end-to-end learning,'' in \emph{Advances in neural
  information processing systems}, 2006, pp. 739--746.

\bibitem{xu2015show}
K.~Xu, J.~Ba, R.~Kiros, K.~Cho, A.~Courville, R.~Salakhudinov, R.~Zemel, and
  Y.~Bengio, ``Show, attend and tell: Neural image caption generation with
  visual attention,'' in \emph{International Conference on Machine Learning},
  2015, pp. 2048--2057.

\bibitem{chen2017brain}
S.~Chen, S.~Zhang, J.~Shang, B.~Chen, and N.~Zheng, ``Brain inspired cognitive
  model with attention for self-driving cars,'' \emph{arXiv preprint
  arXiv:1702.05596}, 2017.

\bibitem{sharmanska2013learning}
V.~Sharmanska, N.~Quadrianto, and C.~H. Lampert, ``Learning to rank using
  privileged information,'' in \emph{Proceedings of the IEEE International
  Conference on Computer Vision}, 2013, pp. 825--832.

\bibitem{jia2014caffe}
Y.~Jia, E.~Shelhamer, J.~Donahue, S.~Karayev, J.~Long, R.~Girshick,
  S.~Guadarrama, and T.~Darrell, ``Caffe: Convolutional architecture for fast
  feature embedding,'' \emph{arXiv preprint arXiv:1408.5093}, 2014.

\bibitem{simonyan2014very}
K.~Simonyan and A.~Zisserman, ``Very deep convolutional networks for
  large-scale image recognition,'' \emph{arXiv preprint arXiv:1409.1556}, 2014.

\bibitem{he2016deep}
K.~He, X.~Zhang, S.~Ren, and J.~Sun, ``Deep residual learning for image
  recognition,'' in \emph{Proceedings of the IEEE conference on computer vision
  and pattern recognition}, 2016, pp. 770--778.

\bibitem{cg23}
``Udacitysdc-challenge2,''
  \url{https://github.com/udacity/self-driving-car/tree/master/steering-models/community-models/cg23},
  accessed: 2016-12-15.

\bibitem{hyndman2008forecasting}
R.~Hyndman, A.~B. Koehler, J.~K. Ord, and R.~D. Snyder, \emph{Forecasting with
  exponential smoothing: the state space approach}.\hskip 1em plus 0.5em minus
  0.4em\relax Springer Science \& Business Media, 2008.

\end{thebibliography}


% \end{thebibliography}

\end{document}